\documentclass[letterpaper, 10pt, conference]{ieeeconf}
\IEEEoverridecommandlockouts
\usepackage{cite}
\usepackage{amsmath,amssymb,amsfonts}
\usepackage{algorithmic}
\usepackage{graphicx}
\usepackage{textcomp}
\usepackage{xcolor}
\usepackage[font=small]{caption}
\usepackage{hyperref} 
\usepackage{booktabs}
\usepackage{array, graphicx, diagbox, tabularx}
\usepackage{fancyhdr}
\usepackage{stfloats}
\usepackage{xcolor}
\usepackage{pifont}
\usepackage{booktabs}
\usepackage[table]{xcolor}
\usepackage{array}
\usepackage{siunitx}

\newcolumntype{C}[1]{>{\centering\arraybackslash}p{#1}}

\newcommand{\cmark}{\textcolor{green!60!black}{\ding{51}}} 
\newcommand{\xmark}{\textcolor{red}{\ding{55}}}

\fancypagestyle{preprint}{
  \fancyhf{}
  \fancyhead[L]{PREPRINT VERSION}
  
  \setlength{\topmargin}{-0.7in}
  \setlength{\headheight}{15pt}
  \setlength{\headsep}{25pt}
}

\newcolumntype{C}{>{\centering\arraybackslash}X}

\def\BibTeX{{\rm B\kern-.05em{\sc i\kern-.025em b}\kern-.08em
    T\kern-.1667em\lower.7ex\hbox{E}\kern-.125emX}}
\begin{document}

\title{The Turkish Ice Cream Robot: \\ Examining Playful Deception in Social Human-Robot Interactions
}
\author{Hyeonseong Kim$^{1}$, Roy El-Helou$^{2}$, Seungbeen Lee$^{3,4}$, Sungjoon Choi$^{1}$ and Matthew Pan$^{2}$%
\thanks{${^1}$Hyeonseong Kim and Sungjoon Choi are with the Department of Artificial Intelligence, Korea University, Seoul, Republic of Korea (e-mails: hyeonseong-kim@korea.ac.kr, chanwoo-kim@korea.ac.kr, sungjoon-choi@korea.ac.kr)}
\thanks{${^2}$Roy El-Helou and Matthew Pan are with the Ingenuity Labs Research Institute and the Department of Electrical and
Computer Engineering, Queen's University, Kingston, Canada (e-mails: 19reh2@queensu.ca, matthew.pan@queensu.ca)}
\thanks{$^{^3}$Seungbeen Lee is with the Department of Artificial Intelligence, Yonsei University, Seoul, Republic of Korea (e-mail: seungblee@yonsei.ac.kr)}
\thanks{$^{^4}$Seungbeen Lee is with the Robotics Institute, Carnegie Mellon University, Pittsburgh, PA, USA (e-mail: seungbel@andrew.cmu.edu)}%
}

\pagestyle{preprint}
\maketitle
\thispagestyle{preprint}

\begin{abstract}
Playful deception, a common feature in human social interactions, remains underexplored in Human-Robot Interaction (HRI). Inspired by the Turkish Ice Cream (TIC) vendor routine, we investigate how bounded, culturally familiar forms of deception influence user trust, enjoyment, and engagement during robotic handovers. We design a robotic manipulator equipped with a custom end-effector and implement five TIC-inspired trick policies that deceptively delay the handover of an ice cream-shaped object. Through a mixed-design user study with 91 participants, we evaluate the effects of playful deception and interaction duration on user experience. Results reveal that TIC-inspired deception significantly enhances enjoyment and engagement, though reduces perceived safety and trust, suggesting a structured trade-off across the multi-dimensional aspects. Our findings demonstrate that playful deception can be a valuable design strategy for interactive robots in entertainment and engagement-focused contexts, while underscoring the importance of deliberate consideration of its complex trade-offs. You can find more information, including demonstration videos, on \url{https://hyeonseong-kim98.github.io/turkish-ice-cream-robot/}.
\end{abstract}

\section{Introduction}
In Human–Robot Interaction (HRI), deceptive behaviours of robots are often treated as harmful since they often reduce user trust~\cite{danaher2020robot, sharkey2021we, saetra2021social, saadon2025scammed}. Prior studies have mainly focused on avoiding deception through predictable behaviours and intent displays~\cite{dragan2013legibility, dragan2014analysis, chen2018planning}, or on repairing trust~\cite{baker2018toward, sebo2019don, rogers2023lying}. On the other hand, empirical investigations into the potential benefits of robot deception have been conducted in game contexts where deception is permitted or even encouraged~\cite{short2010no, de2021deceptive, esposito2025roboleaks}.

However, beyond game-like contexts where the deception is explicitly allowed, some playful deceptions also enrich interpersonal experiences in everyday human-human interaction. Playful deception can create enjoyable, light-hearted moments, so long as it is interpreted as part of a benign performance rather than as manipulation. For instance, a human handing over an object might playfully delay the delivery or briefly mislead the receiver through unexpected but harmless actions, forming humour that is only possible through physical embodiment.

The Turkish ice cream (TIC) vendor routine exemplifies this point: it relies on intentional misdirection to entertain while ultimately delivering the treat. The TIC interaction is a short street performance in which a vendor playfully prolongs the handover of an ice cream cone. Using a long spatula-like rod, the cone is presented and then briefly withdrawn, with light feints and showy gestures that tease the customer while keeping the exchange clearly playful. The routine is recognizably performative and always resolves with a successful handover, marking the deception as benign rather than malicious.

\begin{figure}[t] %
\includegraphics[width=1.0\columnwidth]{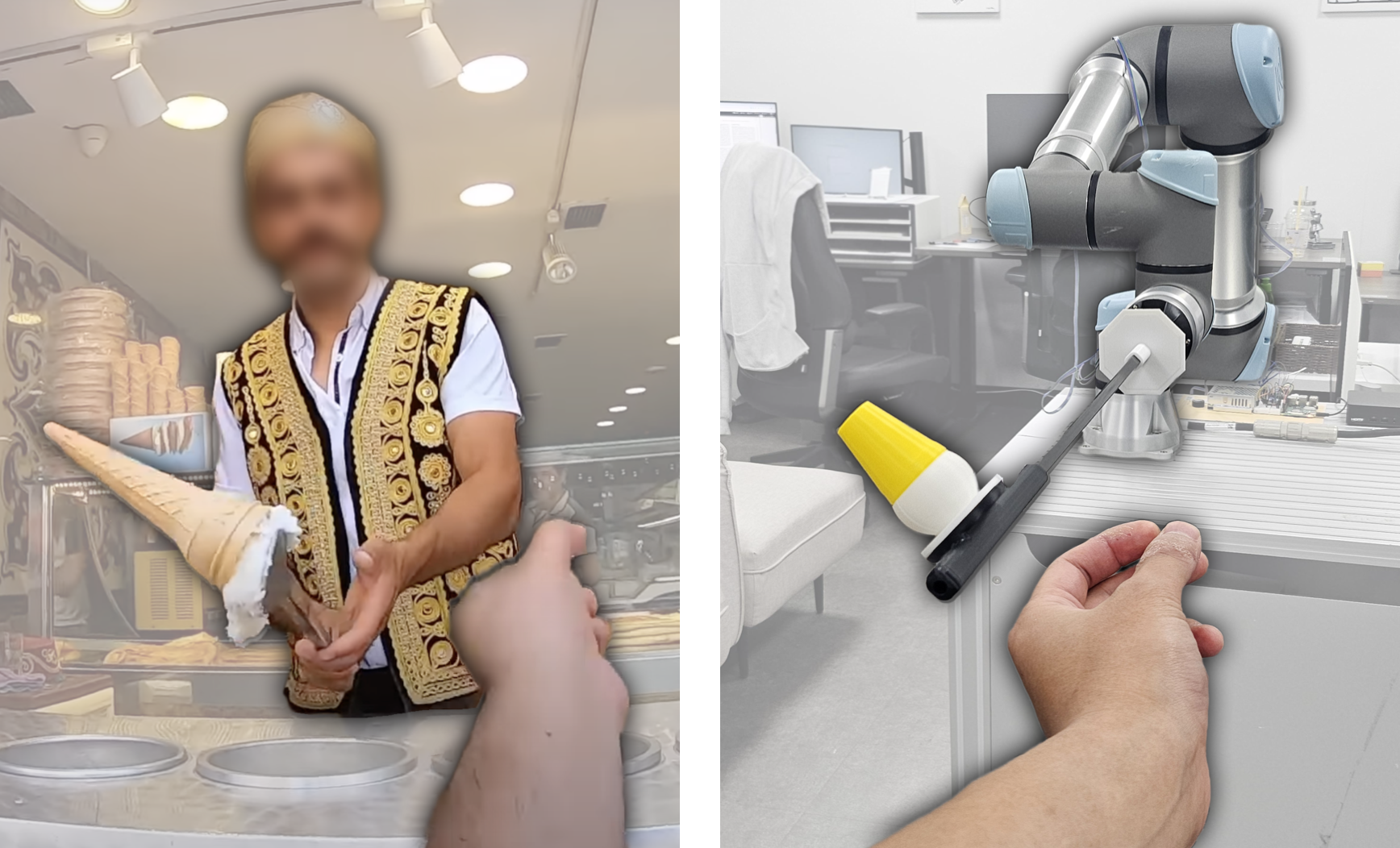}
    \caption{\textbf{Conceptual illustration of the TIC robot.} Inspired by the playful techniques of Turkish ice cream vendors~\cite{tic_concept} (left), the robot is designed to humorously mislead the user while handing over an ice cream cone-shaped object (right).}
    \label{fig:concept}
    \vspace{-7pt}
\end{figure}

In this paper, we explore the potential role of playful deception as a design component in HRI through a study inspired by the TIC routine. To this end, as shown in Fig.~\ref{fig:concept}, we design and implement a robotic handover system that reproduces TIC-inspired playful deceptive behaviours and conduct a user study with 91 participants to investigate the multi-dimensional effects of playful deception, such as enjoyment, trust, and perceived safety.

In summary, this paper makes three primary contributions:
\begin{itemize}
    \item We design and implement a robotic system capable of reproducing TIC-inspired deceptive handovers, introducing a novel playful interaction scenario.
    \item We conduct a large-scale user study evaluating the multi-dimensional effects of playful TIC deception on user experience, revealing both its positive and negative impacts.
    \item We emphasize the nuanced trade-offs that playful deception introduces and the importance of situating its use within appropriate contexts, offering guidance for both design and future research.
\end{itemize}

\section{Backgrounds}
Robot deception in HRIs has often been seen as problematic because it can lead to undermining user trust and ethical concerns. Prior studies have focused on defining forms and analyzing the effect of robot deception~\cite{danaher2020robot, sharkey2021we, saetra2021social, saadon2025scammed}, designing motion strategies to avoid unintended misunderstandings~\cite{dragan2013legibility, dragan2014analysis, chen2018planning}, and developing methods to recover trust when it is damaged~\cite{baker2018toward, sebo2019don, rogers2023lying}. While these approaches generally assume that deception is inherently harmful, \cite{wagner2011acting, shim2013taxonomy, esposito2025deception} suggest the potential benefit of robot deception in HRI scenarios. Some work has empirically shown that deception can improve engagement of the user and the perception about the robot's competence~\cite{short2010no, de2021deceptive, esposito2025roboleaks}, but such findings are mostly limited to game settings where explicit rules allow deceptive strategies. Although a few studies~\cite{shim2016other} examined the multi-dimensional effects of benevolent deception in contexts such as patient and elderly care, its potential role in everyday HRI scenarios remains largely unexplored.

In this paper, we hypothesize that \emph{playful deception}—when designed to be socially acceptable and safely bounded—may also have multi-dimensional effects in everyday interaction scenarios. In other words, rather than treating deception purely as a risk to be avoided, it may serve as a potential interaction design component under certain conditions. This expectation is grounded in two theoretical frameworks from psychology: incongruity theory~\cite{forabosco2008concept}, which posits that humour arises when expectations are disrupted, and benign violation theory~\cite{mcgraw2010benign}, which suggests that norm violations can be experienced humorously when framed as safe and acceptable. Together, these theories motivate the idea that carefully designed playful deception can elicit enjoyment, operating as an interaction strategy.

Based on these insights, we formulate two main hypotheses:
\begin{itemize}
\item \textbf{H1.} Playful, safety-bound deception can increase users’ enjoyment and engagement but may decrease trust and perceived safety.
\item \textbf{H2.} The effects of playful deception can be moderated by interaction design factors, such as motion style, timing, and sequence.
\end{itemize}

However, identifying or defining a scenario that naturally represents socially acceptable and safely bounded deception in everyday interactions is challenging. To address this, we draw inspiration from the Turkish ice cream vendor routine, a human-human interaction that relies on playful misdirection while maintaining clear safety and intent boundaries. The TIC scenario provides a natural, culturally recognizable setting where playful deception enhances the experience rather than undermines it.

Thus, the goal of this research is to design and implement a Turkish ice cream robot capable of reproducing TIC-inspired deceptive behaviours. Utilizing this system, we explore the potential role of playful deception as a design element in HRI by analyzing its multi-dimensional effects on user experience and observing how the deceptive interaction design parameter, especially the interactivity duration, can influence those effects.

\section{The Turkish Ice Cream Robot}

\subsection{Overview and Design Objectives}
\begin{figure*}[!t] %
\includegraphics[width=2.05\columnwidth]{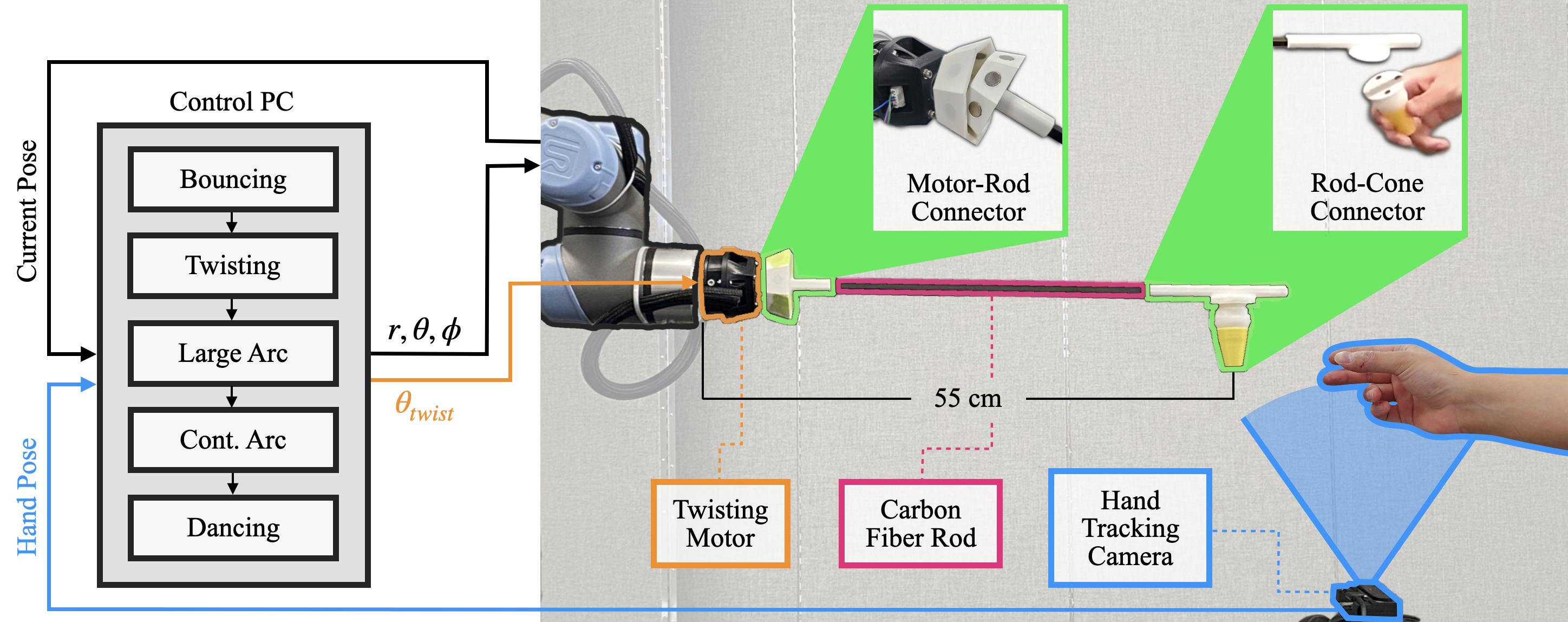}
    \caption{\textbf{System overview.} The system takes the participant’s hand pose and the robot’s current pose as inputs. Five trick policies are deployed sequentially, each executed for a predefined interactivity duration. Based on the current policy, the controller computes $(r, \theta, \phi)$ for the manipulator and $\theta_{\mathrm{twist}}$ for the twisting motor. A rod-shaped end-effector amplifies cone motion, while the magnetic connectors (highlighted in green) enable safe and seamless handovers.}
    \label{fig:system_overview}
    \vspace{-7pt}
\end{figure*}

Our goal in this work is to investigate how playful deception inspired by the TIC routine can affect user experience in HRIs. To achieve this, we have designed a robotic handover system that can reproduce key elements of the TIC routine while ensuring user safety and experimental control.

Specifically, we set the following design objectives:
\begin{enumerate}
    \item 
    Implement a hardware module that ensures safe interaction even in unintended collisions.
    \item 
    Capture and reproduce the characteristic motions of TIC vendors while ensuring the deception remains socially acceptable and playful.
    \item 
    Implement multiple trick behaviours derived from observational analyses, allowing controlled manipulation of deceptive strategies and interaction time.
\end{enumerate}

With these objectives, we built a \emph{custom TIC End-effector module} that allows the robot to achieve swift and safe evasive motions (Section~\ref{sec:end-effector}), 
developed \emph{TIC trick policies} to reproduce key deceptive strategies observed of TIC vendors (Section~\ref{sec:policies}), 
and integrated \emph{implementation details}, including inverse kinematics, hand tracking, and system synchronization into a unified framework (Section~\ref{sec:details}). Figure~\ref{fig:system_overview} illustrates an overview of the entire system.

\subsection{Custom TIC End-Effector Module} \label{sec:end-effector}
To replicate the equipment and characteristic gestures of TIC vendors, we developed a custom end-effector module that mounts to the manipulator and extends its reach. The module consists of a \textit{twisting motor}, a \textit{rod}, and a \textit{cone} handover object (Fig.~\ref{fig:system_overview}). Placing the cone at the rod tip not only reproduces the vendor’s tool but also amplifies the manipulator’s motion: small joint movements translate into larger, faster displacements, enabling expressive evasive gestures without requiring extreme joint speeds.

\subsubsection{Twisting Motor}
Because the manipulator’s wrist joints are too slow and limited in range to reproduce TIC-style tricks, we employ a stepper motor to generate rapid twisting gestures. The motor receives the target rotation angle $\theta_{twist}$, is mounted directly to the manipulator, and is coupled to the rod via a custom connector.

\subsubsection{Rod}
A carbon-fibre rod is attached to the motor by neodymium magnets, reinforced with masking tape for stability during high-speed motion. This hybrid attachment provides rigidity while allowing safe detachment in case of unintended contact. Measuring 55~cm, the rod extends the manipulator’s reach: subtle wrist rotations produce large, agile displacements at the cone tip, increasing dodging speed without requiring large joint excursions.

\subsubsection{Cone}
The cone is magnetically mounted at the rod tip, allowing users to easily detach and retrieve it. To mimic the appearance of ice cream, it is divided into a white “ice cream” portion and a yellow “cone” portion. Participants were instructed to grasp only the yellow section, which reduced the graspable area, increased capture difficulty, and reinforced the playful misdirection in TIC interactions.

\subsection{TIC Trick Policies} \label{sec:policies}
The original TIC performance involves a wide range of playful misdirections, including cone switching, conversational distractions, and gaze-based deception. However, our robotic platform is limited in both sensing and actuation modalities, making it infeasible to replicate the full richness of human vendor behaviours. To ensure both feasibility and safety, we focus on a subset of TIC behaviours where deception primarily arises from evasive physical motions.

\subsubsection{Extracting Features from Human TIC Interactions}
To design these motions, we conduct an observational analysis of TIC performance videos found online and extract key performance features that contribute to the playful nature of the tricks. From this analysis, we extracted four recurring features that capture essential aspects of the vendor’s movement strategies: \textit{least effort}, \textit{near-miss}, \textit{misdirection}, and \textit{exaggeration}.

\begin{itemize}
    \item \textbf{Least Effort} refers to the vendor’s use of minimal physical motion to maintain control of the cone while still preventing the user from grasping it. For example, the vendor may subtly retract the cone by just a few centimetres as the customer reaches forward, using the wrist only. This aligns with Zipf’s principle of least effort \cite{zipf2016human} and is often employed to sustain user attention efficiently.

    \item \textbf{Near-Miss} describes situations where the customer's hand comes very close to grasping the cone but fails by a small margin. One common example is when the vendor quickly flips the cone upside down just before contact. This near-success amplifies engagement by introducing momentary tension and surprise, a mechanism also discussed in gambling theories \cite{reid1986psychology}.

    \item \textbf{Misdirection} involves deliberately diverting the customer’s attention from the actual movement of the cone. For instance, the vendor may use one hand to draw attention while smoothly switching the cone to the other, taking advantage of the user’s focus being elsewhere. This technique, also used in magic \cite{macknik2010sleights}, helps sustain unpredictability.

    \item \textbf{Exaggeration} involves amplified or overly dramatic gestures that emphasize the motion of the cone or the body of the vendor. A typical example includes twirling the cone in large arcs above the user’s head. Such gestures enhance expressiveness and create a playful rhythm in the interaction.
\end{itemize}

These four features serve as the conceptual foundation for our robot’s behaviour design, guiding both the structure of each trick and the dynamics of timing, effort, and attention within the interaction flow.

\begin{figure*}[!t] %
    \centering
    \includegraphics[width=2.05\columnwidth]{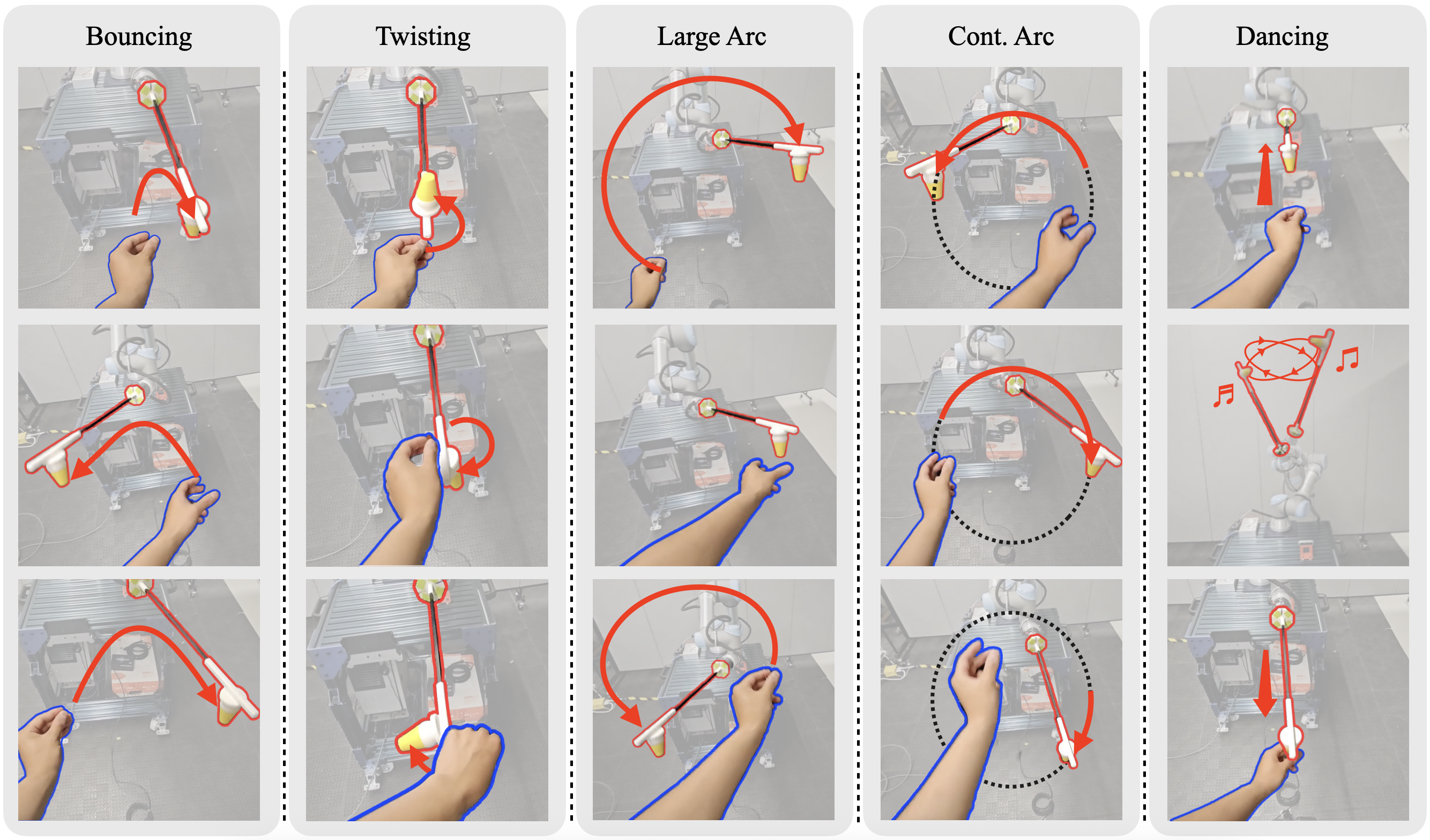}
    \caption{\textbf{Visualization of the five distinct trick motion policies.} Each column represents a different policy, showing the corresponding motion trajectories executed during real-world demonstrations. From top to bottom, each row illustrates the temporal progression of the motion within each policy.}
    \label{fig:policies}
    \vspace{-7pt}
\end{figure*}

\subsubsection{Trick Behaviour Policies}
Based on the extracted features of TIC tricks, we design five distinct trick behaviour policies. As shown in Tab.~\ref{tab:policy_feature}, each policy (rows) has one or more characteristics of the TIC vendor trick (columns). Employing multiple policies allows for variation in the robot’s motion, making it more difficult for the user to anticipate the next movement. This unpredictability increases the likelihood of failed grasp attempts, thereby sustaining engagement and reinforcing the playful nature of the interaction. Fig.~\ref{fig:policies} visually describes the motion of each policy.

\begin{table}[!b]
    \renewcommand{\arraystretch}{1.1} 
    \caption{Policy-characteristic mapping of implemented TIC tricks.}
    \begin{tabularx}{\columnwidth}{@{} lcccc @{}} 
        \toprule
        \textbf{Policy} & \textbf{Least Effort} & \textbf{Near-Miss} & \textbf{Misdirection} & \textbf{Exaggeration} \\
        \midrule
        Bouncing      & \cmark & \xmark & \cmark & \xmark \\ 
        Twisting      & \cmark & \cmark & \xmark & \xmark \\ 
        Large Arc     & \xmark & \cmark & \cmark & \cmark \\ 
        Cont. Arc    & \xmark & \xmark & \cmark & \xmark \\ 
        Dancing       & \xmark & \cmark & \xmark & \cmark \\ 
        \bottomrule
    \end{tabularx}
    \label{tab:policy_feature}
\end{table}

\begin{itemize}
    \item \textbf{Bouncing} makes the robot move the cone side-to-side in a rhythmic, hopping motion. Each bounce traces a semicircular trajectory, creating the illusion that the cone is "hopping" away from the user. As the arc diameter increases with each bounce, the user must cover more distance to follow. Simultaneously, the arc motion reduces collision risk by discouraging direct linear hand movements toward the target.

    \item \textbf{Twisting} rotates the cone to the opposite side of the approaching hand while the shaft position is fixed at a certain position. This is one of the fundamental policies that real TIC vendors utilize, where the vendors bring out customers' large action, spending little energy. 

    \item \textbf{Large Arc} is the discrete action policy that dodges the hand by drawing a large arc path when the hand reaches the cone. The cone escapes from the hand narrowly and then draws an exaggerated, huge arc. Drawing a large arc has two advantages: it avoids collision with the hand while the robot moves, and it makes the user confused about where the destination of the cone is.

    \item \textbf{Continuous Arc (Cont. Arc)} controls the robot to draw an arc whose center is fixed to a certain position. The robot moves to the opposite side of the hand on the circle continuously. This movement looks like a magnetic repulsive force and misdirects the user to follow the cone in a circular path, while the most effective movement of the user for reaching the cone is just to go straight to the cone.

    \item \textbf{Dancing} is a non-interactive policy that is intended to tease/taunt users by circularly waving the rod upright out of the reachable range of the users. The robot moves the cone backward when the hand reaches the cone and makes a circle while twisting the cone. At the end of the policy, the robot extends the cone toward the user, finishing the interaction.
\end{itemize}

\subsection{Implementation Details} \label{sec:details}

\subsubsection{Robot and Inverse Kinematics}
We use a UR5e manipulator \cite{ur5e} to execute the TIC robot’s motions. The 6-DOF arm is controlled in a reduced 3-DOF task space anchored to a human-centered frame, parameterized as $(r,\theta,\phi)$: a prismatic degree of freedom $r$ along the forward--back (approach) axis, and two angular degrees of freedom $\theta$ (azimuth) and $\phi$ (elevation) that sweep the end effector laterally and vertically on a sphere of fixed radius. This spherical parameterization simplifies planning without compromising interaction quality, while reflecting real TIC routines where vendors achieve large, expressive motions with subtle wrist movements. Joint commands are computed via inverse kinematics, subject to joint and velocity constraints to ensure smooth, safe execution.

\subsubsection{Hand Tracking}
We use a Leap Motion Controller 2 \cite{leapmotion} to track participants’ hand pose at 120~Hz. The sensor is positioned below the hand, facing upward, to ensure reliable detection while minimizing collision risk (Fig.~\ref{fig:system_overview}, right). Participants grasp the cone using a pinching gesture, with the pinch position---defined as the midpoint between the thumb and index fingertip---serving as the target grasp point for its stability and precision. To mitigate latency effects on interaction timing, we apply a velocity-based prediction with a 0.15~s lookahead and a low-pass filter to smooth noisy measurements.

\subsubsection{System Integration}
All sensory and actuation data, including the hand pose, twisting motor pose, and UR5e end-effector pose, are collected and synchronized in real time through the ROS2 framework. The UR5e manipulator is controlled at 200~Hz to ensure smooth and stable handover trajectories, while the twisting motor operates at 50~Hz to reproduce fast and precise rod rotations.
\section{User Study}

\subsection{Research Questions}
We propose three research questions to test our two hypotheses. To operationalize H1, we formulated one research question: 
(\textbf{RQ1}) How do perceptions of the robot differ between straight handover and deceptive handover?
To operationalize H2, we formulated a research question: 
(\textbf{RQ2}) How does interaction length (short, medium, long) influence user perceptions of deceptive handovers?

\subsection{Procedure}
To evaluate the effects of playful deception and interaction timing in physical handovers, we conducted a mixed-design user study. All participants experienced two handover types: a \textbf{Straight Handover} (\textbf{SH}, baseline) and a \textbf{Deceptive Handover (DH)}, presented in randomized order to counterbalance potential ordering effects. Participants were told that the study involved robot-to-human handover, but no mention was made of deception or TIC-inspired interaction.

\begin{figure*}[!t] %
    \centering
    \includegraphics[width=2.05\columnwidth]{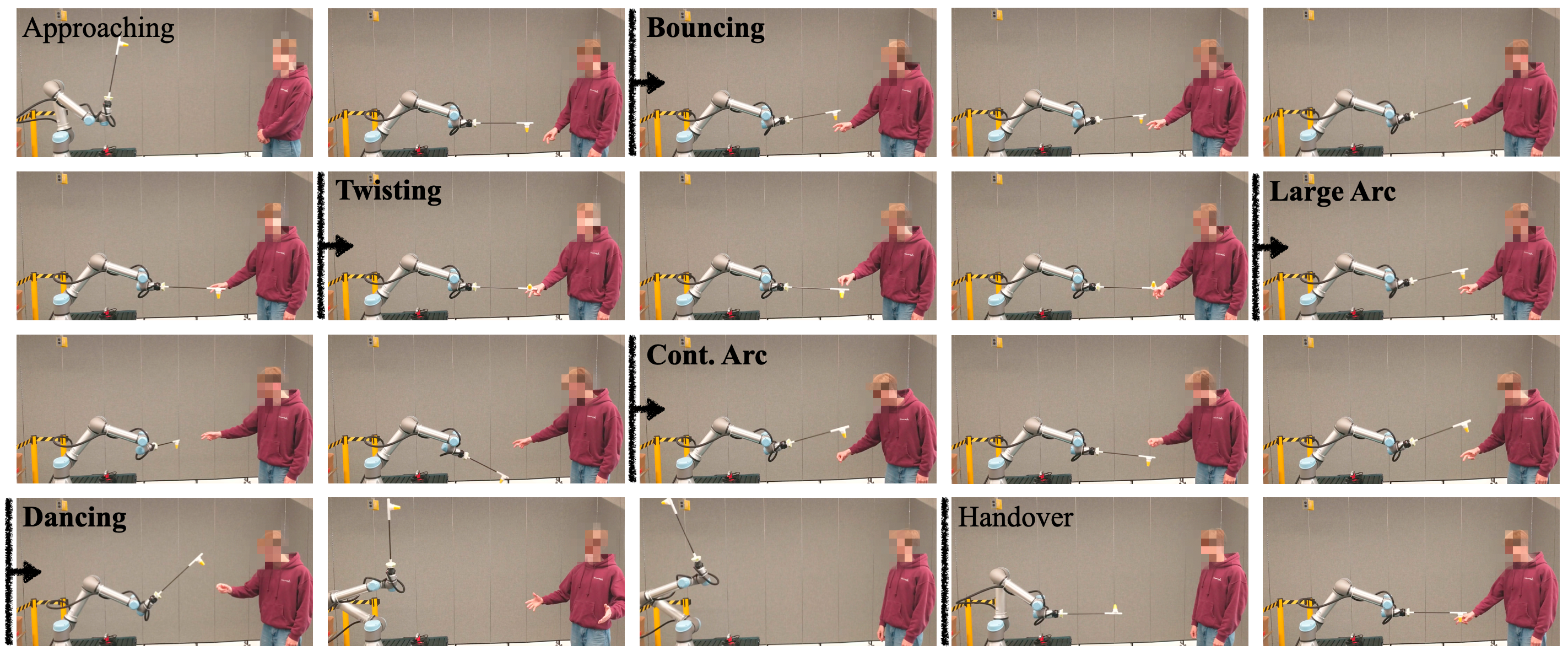}
    \caption{\textbf{User study snapshots of the Deceptive Handover (DH).} The interaction proceeds from top left to bottom right, beginning with approaching (first two frames) and then executing five trick policies sequentially: \textit{Bouncing}, \textit{Twisting}, \textit{Large Arc}, \textit{Cont. Arc}, and \textit{Dancing}. The bold-named frames mark the first image of each policy for clarity. Each policy was performed for its pre-assigned interaction duration. At the end, the cone approaches again to give the cone (last two frames), while the Straight Handover (SH) condition simply performs this final handover without deception. The corresponding motion sequences can be found in the supplementary video.}
    \label{fig:user_study_snapshot}
    \vspace{-7pt}
\end{figure*}

In the \textbf{SH} condition, the robot handed the cone directly to the user without any deceptive motions. The cone was placed slightly to the left or right of the user’s midline, depending on which hand the user extended, based on the prior finding that users generally prefer handovers to occur slightly on the receiver's reaching hand's side~\cite{basili2009investigating}.

The \textbf{DH} condition was divided into three sub-conditions based on interactivity duration: \textit{Short}, \textit{Medium}, and \textit{Long}. The robot executed a fixed sequence of five tricks in the following order: \textit{bouncing (4 reps)}, \textit{twisting (3s)}, \textit{large arc (4 reps)}, \textit{cont. arc (3s)}, and \textit{dancing (3s)}. These parameters defined the \textit{Medium} condition, which served as the baseline. The \textit{Short} and \textit{Long} conditions were created by scaling all durations and repetitions by factors of 0.5 and 2, respectively, while keeping the behavioural sequence unchanged. This manipulation systematically varied interaction time while maintaining consistency in motion patterns across participants. Fig.~\ref{fig:user_study_snapshot} illustrates the snapshots of the DH condition in the user study setting.

Participants were randomly assigned to one of the three duration conditions in a between-subjects design, while the comparison between SH and DH was evaluated within-subjects. Each participant took approximately 30 minutes to complete the experiment. 

\subsection{Measures}
We used a combination of standardized scales and custom questions to assess participants’ perceptions, trust, enjoyment, and qualitative impressions. After each condition, they completed two questionnaires based on 7-point Likert scales for each inventory item (1 - strongly disagree, 7 - strongly agree): the Multi-Dimensional Measure of Trust (MDMT) v2 \cite{malle2021multidimensional}, which measures reliability, competence, ethics, transparency, and benevolence; and the ENJOY scale \cite{davidson2023development}, which measures pleasure, self-competence, challenge, and engagement. Participants also reported rated perceived deception on a 7-point Likert scale, their willingness-to-pay (WTP) for an ice cream served by the robot.

\section{Results}
A total of 91 participants were recruited to participate without compensation through on- and off-campus advertising—one participant's data was not recorded due to equipment failure and was not reported here. Participants self-identified as men (n=54), women (n=35), and non-binary (n=1). No other gender identities were a part of the sample. Ages of the participants ranged from 18 to 66 years old ($M = 27.9$, $SD = 10.5$). This study was approved by the Queen’s University General Research Ethics Board (GELEC-139-22, File No. 6036728).

\subsection{Analysis}
We conducted mixed-design repeated-measures ANOVAs (RM-ANOVAs) to examine changes in perceptions from SH to DH and whether these depended on interactivity duration (short, medium, long). \textbf{Handover Type (SH vs. DH)} served as the within-subjects factor and \textbf{Interactivity Duration} as the between-subjects factor, allowing us to test both overall SH–DH differences and their interaction with length. Estimated marginal means with Bonferroni-adjusted comparisons clarify effect directions. To assess interactivity duration independently, one-way ANOVAs were run on DH scores alone. This design was appropriate because each participant experienced one SH and only one DH. Effect sizes are reported as partial eta squared ($\eta^2$), with 0.01, 0.06, and 0.14 denoting small, medium, and large effects, respectively \cite{cohen2013statistical}.

\begin{figure*}[!t]
    \centering
    \includegraphics[width=2.05\columnwidth]{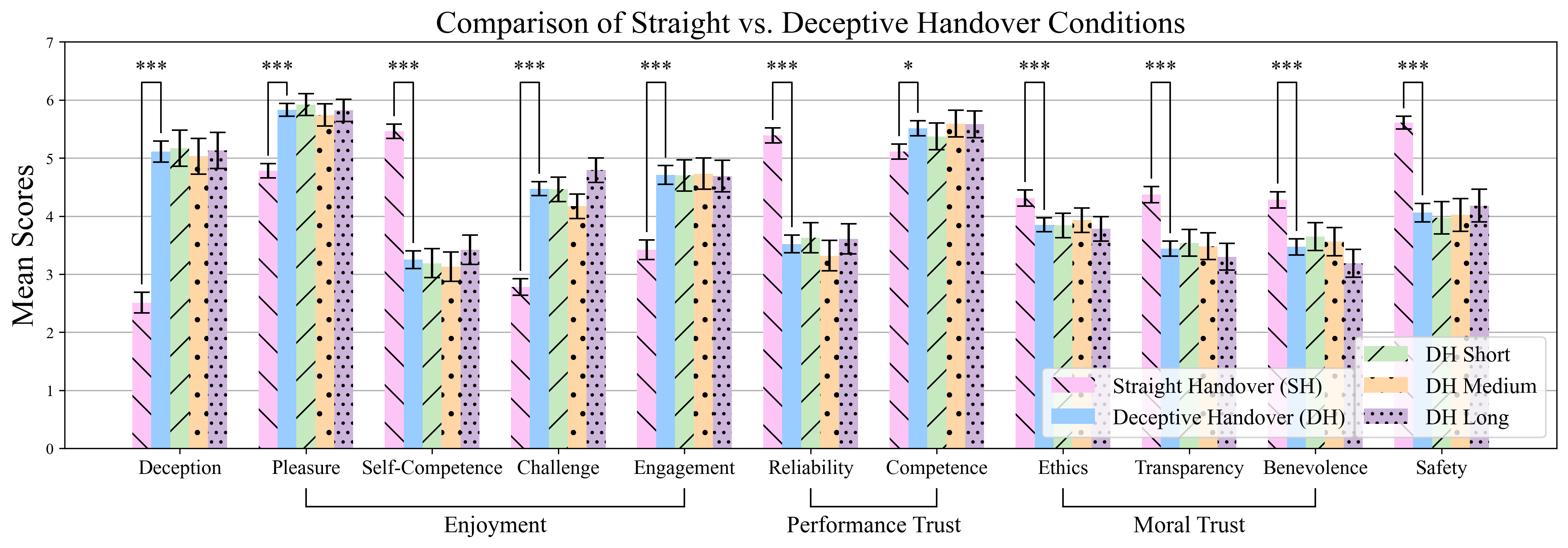}
    \caption{\textbf{User study results comparing Straight (SH) vs. Deceptive Handover (DH).} DH Short, Medium, and Long indicate variations in interactivity duration, and DH represents the combined results across all durations. DH was perceived as more deceptive than SH, while increasing enjoyment-related outcomes (pleasure, challenge, engagement) and competence, but reducing trust, safety, and self-competence. Error bars represent $\pm$Standard Error (SE) and asterisks denote significance levels with Bonferroni correction: \textsuperscript{*}$p < .05$, \textsuperscript{***}$p < .001$.}
    \label{fig:user_study}
\end{figure*}

\subsection{Straight vs. Deceptive Handover Main Effects (RQ1)}
To address RQ1, we tested whether perceptions differed between Straight (SH) and Deceptive Handover (DH). The RM-ANOVA showed a significant overall effect of Handover Type, $F(12,76)=27.31$, $p<.001$, $\eta^2=.812$.  

As summarized in Fig.~\ref{fig:user_study}, participants rated DH as more deceptive than SH, confirming the manipulation. DH also increased enjoyment-related outcomes (pleasure, engagement, challenge) and perceived robot competence, but decreased performance trust (reliability), moral trust (ethics, transparency, benevolence), perceived safety, and self-competence. Bonferroni-corrected pairwise comparisons indicated these effects were consistent across most measures ($p<.05$). Full results are reported in Tab.~\ref{tab:sh_dh}.

\begin{table}[!t]
\centering
\caption{SH vs. DH ANOVA result details. Significant effects ($p < .05$) are shown with corresponding effect sizes ($\eta^2$).}
\label{tab:sh_dh}
\renewcommand{\arraystretch}{1.25}
\begin{tabularx}{\linewidth}{
  l
  S[table-format=3.2]  
  c                    
  S[table-format=1.2]  
  C                    
}
\toprule
\textbf{Measure} & $\boldsymbol{F(1,87)}$ & {$\boldsymbol{p}$} & {$\boldsymbol{\eta^2}$} & \textbf{Effect Direction} \\
\midrule
\rowcolor{gray!12}\multicolumn{5}{l}{\textbf{Deception}} \\
Deception & 92.28 & $<.001$ & .52 & DH $>$ SH \\
\midrule
\rowcolor{gray!12}\multicolumn{5}{l}{\textbf{ENJOY (Enjoyment)}} \\
Pleasure        & 70.75  & $<.001$ & .45 & DH $>$ SH \\
Self-Competence & 139.16 & $<.001$ & .62 & SH $>$ DH \\
Challenge       & 110.03 & $<.001$ & .56 & DH $>$ SH \\
Engagement      & 63.59  & $<.001$ & .42 & DH $>$ SH \\
\midrule
\rowcolor{gray!12}\multicolumn{5}{l}{\textbf{MDMT (Performance Trust)}} \\
Reliability     & 87.25  & $<.001$ & .50 & SH $>$ DH \\
Competence      & 5.82   & $<.05$  & .06 & DH $>$ SH \\
\midrule
\rowcolor{gray!12}\multicolumn{5}{l}{\textbf{MDMT (Moral Trust)}} \\
Ethics          & 14.77  & $<.001$ & .15 & SH $>$ DH \\
Transparency    & 47.84  & $<.001$ & .36 & SH $>$ DH \\
Benevolence     & 31.93  & $<.001$ & .27 & SH $>$ DH \\
\midrule
\rowcolor{gray!12}\multicolumn{5}{l}{\textbf{Godspeed (Perceived Safety)}} \\
Perceived Safety & 86.29  & $<.001$ & .50 & SH $>$ DH \\
\bottomrule
\end{tabularx}
\vspace{-7pt}
\end{table}

\subsection{Effects of Interaction Durations Across in DH (RQ2)}
To address RQ2, we examined whether interactivity duration influenced participants’ perceptions of deceptive handovers. As shown in Fig.~\ref{fig:user_study}, the direction of effects is consistent across measures for SH vs. DH as compared to SH vs. any DH duration. However, their levels of significance do differ: competence is non-significant across all durations. Ethics is non-significant for the short duration but significant at $p < .05$ for the long duration. Transparency ($p < .01$) and benevolence ($p < .05$) are significant for the short duration, while pleasure is significant at $p < .01$ for the moderate duration. All other measures and durations not mentioned are significant at the $p < .001$ level. Furthermore, within the DH condition itself, no significant differences emerged between the three duration groups across any of the 13 measures.

\subsection{Willingness to Pay}
WTP was reported by participants in CAD, but has been converted to US dollars.\footnote{We use a CAD $\rightarrow$ USD exchange rate of 0.72} The RM-ANOVA showed no overall difference between SH and DH. For the short interaction group, WTP did not significantly differ between SH 
($M=\$8.06$~USD, $SE=\$2.53$~USD) 
and DH 
($M=\$6.90$~USD, $SE=\$2.31$~USD), 
$p=.58$. 
Similarly, for the moderate interaction group, no significant difference was found between SH 
($M=\$4.51$~USD, $SE=\$2.39$~USD) 
and DH 
($M=\$5.39$~USD, $SE=\$2.19$~USD), 
$p=.66$. In the long interaction group, WTP was significantly higher for DH 
($M=\$9.87$~USD, $SE=\$2.20$~USD) 
than SH 
($M=\$4.59$~USD, $SE=\$2.04$~USD),
$p=.01$. This suggests that while deception does not consistently affect WTP across interaction durations, participants in the longest condition expressed greater willingness to pay. A one-way ANOVA across the three DH duration conditions revealed no significant differences.

\subsection{Retrieval Success vs. Perceived Safety}
We tested whether participants’ success in retrieving the cone during the DH condition influenced perceived safety. Retrieval success (Successful, S vs. Non-Successful, NS) was treated as a between-subjects factor, with perceived safety ratings as the dependent variable. An independent-samples $t$-test showed no significant difference between participants who retrieved the cone ($S$, $M=3.90$, $SD=1.36$) and those who did not ($NS$, $M=4.19$, $SD=1.60$), $t(88)=0.75$, $p>.05$. 

\section{Discussion}
The study investigated how playful, TIC-inspired deception influences user experience in robot handovers. The results support H1; deceptive handovers increased enjoyment-related measures but reduced multiple dimensions of trust and perceived safety. Specifically, participants rated the deceptive handover (DH) condition as significantly higher in pleasure, challenge, and engagement (ENJOY subscales) compared to the straight handover (SH), while also perceiving the robot as more competent. At the same time, DH scored significantly lower on performance trust (reliability), moral trust (ethics, transparency, benevolence), perceived safety, and self-competence, indicating that playful deception created a clear trade-off between enjoyment and predictability. 

With respect to H2, the results provided limited support. We hypothesized that interaction design factors, such as interaction duration, might moderate the effects of deception. Although several measures' significance level between SH and DH varied across the interactivity durations, there were no systematic shifts in the overall pattern: the tendency observed in SH vs. DH---greater enjoyment and engagement but lower trust and perceived safety in DH---remained stable across short, medium, and long durations. Moreover, comparisons within the DH conditions themselves revealed no significant differences, suggesting that user perceptions were shaped more by the presence of deceptive behaviour than by its duration.

The WTP results add nuance to these findings; while deception did not affect WTP overall, participants in the long interaction group reported significantly higher willingness to pay for the deceptive handover. This suggests that prolonged playful deception may increase the perceived entertainment value of the interaction. However, given the high degree of variability in responses, this effect should be interpreted cautiously. 

In addition, participants’ perception of safety was not contingent on whether they successfully retrieved the cone. Instead, safety judgments appear to have been shaped more by the robot’s deceptive behaviour itself rather than the functional outcome of the interaction. This highlights that, in deceptive HRI scenarios, perceived safety is less about task success and more about the predictability and transparency of robot behavior. This result underscores the importance of considering how interaction style---not just performance outcomes---shapes user trust and comfort in social robots.

The findings contribute to three broader insights for HRI. First, they demonstrate that deception is not inherently harmful; when bounded and framed as playful, it can reliably enhance enjoyment and sustain engagement. This aligns with earlier work showing that deceptive strategies can heighten engagement in game-like HRI contexts \cite{shim2013taxonomy, de2021deceptive}, but extends those insights into embodied, physical handovers, a more everyday form of social interaction. 

Second, the results reinforce that the deception's value is context-dependent. In domains where safety, transparency, and reliability are paramount (e.g., healthcare, industrial collaboration, transportation), the observed decreases in trust and perceived safety would likely undermine acceptance. However, in entertainment, hospitality, or retail, playful misdirection could be leveraged as a design feature. Much like the Turkish ice cream vendor routine, a robot that temporarily frustrates before delighting may create more memorable and socially engaging experiences. 

Third, the asymmetry between increased perceived robot competence and decreased user self-competence highlights that competence in HRI is relational rather than absolute. The robot's smooth, controlled tricks were interpreted as skillful, while simultaneously undermining the participants' sense of their own effectiveness. This echoes work on trust violation and repair \cite{rogers2023lying}, raising questions about how repeated exposure to playful deception might impact long-term user confidence and willingness to collaborate. 

Finally, the results are consistent with benign violation theory \cite{mcgraw2010benign}, breaking expectations in a safe and bounded way produced positive surprise and enjoyment, but also introduced uncertainty about predictability and intent, which reduced trust and safety. For HRI design, at least in carefully structured scenarios, this means playful deception should be deployed selectively so that the benefits of entertainment and engagement do not come at unacceptable costs. 

\section{Limitations and Future Work}
This study has several limitations that shape the scope of its conclusions. First, the deceptive behaviours were implemented as a fixed set of justified trick policies. While these captured key features of TIC-style tricks, they may lack the natural timing and variability of adaptive performances, which can be additional design parameters rather than the interactivity duration. Second, the participants experienced only a single deceptive and straight handover in a laboratory setting. As such, the results capture immediate, one-shot impressions rather than long-term dynamics; repeated exposure could either diminish novelty or erode trust more severely. Finally, the measure relied primarily on self-report questionnaires, and although they revealed clear patterns, complementary behavioural and physiological data could provide stronger evidence about the processes underlying engagement, safety perception, and trust. 

Taken together, these limitations mean the results can be trusted as reliable evidence of how scripted, short-term playful deception affects immediate perceptions in controlled settings, but they should not be assumed to generalize fully to long-term or applied contexts. Future work should address these limitations by testing a broader range of design parameters, conducting longitudinal studies, and exploring field deployments to further complete our understanding of deception in HRI. 

\section{Conclusion}

This work examined whether playful, TIC-inspired deception changes user experience in robot handovers and whether these effects depend on interactivity durations. We compared straight and deceptive handovers, varying interactivity durations across short, moderate, and long conditions. The results show that deceptive handovers consistently increased enjoyment, engagement, and perceived robot competence, while reducing performance trust, moral trust, perceived safety, and user self-competence. These effects were robust across interaction durations, indicating that the presence of deception itself, rather than its duration, primarily shaped participants' perceptions. Thus, our findings answer the first research question directly, while providing only limited support for moderation by timing. 

The key takeaway is that playful deception produces a structured tradeoff: it can delight and sustain attention but at the cost of predictability and trust. This positions deception not as inherently good or bad, but as a contextual design choice. In domains prioritizing entertainment or memorability, bounded misdirection may be valuable. In safety-critical applications, however, the associated declines in trust and safety would likely be unacceptable. These conclusions must be viewed in light of study limitations, scripted behaviours, single-trial laboratory settings, and reliance on self-report. Future work should explore more varied design parameters, repeated interactions, and field studies. Overall, our findings highlight deception as a multi-dimensional tool, one that, if carefully bounded, can balance engagement and trust in human-robot interaction. 


\bibliographystyle{IEEEtran}
\bibliography{references}
\end{document}